# A Modified AUC for Training Convolutional Neural Networks: Taking Confidence into Account

Khashayar Namdar[1,2], Masoom A. Haider[3,4], Farzad Khalvati[1,2]

[1] Department of Medical Imaging, University of Toronto. Toronto, ON, Canada
[2] The Hospital for Sick Children (SickKids), Toronto, ON, Canada
[3] Lunenfeld-Tanenbaum Research Institute, Sinai Health System, Toronto, ON, Canada
[4] Sunnybrook Research Institute, Toronto, ON, Canada
ernest.namdar@utoronto.ca

**Abstract.** Receiver operating characteristic (ROC) curve is an informative tool in binary classification and Area Under ROC Curve (AUC) is a popular metric for reporting performance of binary classifiers. In this paper, first we present a comprehensive review of ROC curve and AUC metric. Next, we propose a modified version of AUC that takes confidence of the model into account and at the same time, incorporates AUC into Binary Cross Entropy (BCE) loss used for training a Convolutional neural Network for classification tasks. We demonstrate this on three datasets: MNIST, prostate MRI, and brain MRI. Furthermore, we have published GenuineAI, a new python library, which provides the functions for conventional AUC and the proposed modified AUC along with metrics including sensitivity, specificity, recall, precision, and F1 for each point of the ROC curve.

## Introduction

Classification is an important task in different fields, including Engineering, Social Science, and Medical Science. To evaluate quality of classification, a metric is needed. Accuracy, precision, and F1 score are three popular examples. However, there are other metrics that are more accepted in specific fields. For example, sensitivity and specificity are widely used in Medical Science.

For binary classification, Receiver Operating Characteristic (ROC) curve incorporates different evaluation metrics. The Area Under ROC Curve (AUC) is a widespread metric, especially in Medical Science [1]. In engineering, AUC has been used to evaluate the classification models since the early 1990s [2], and AUC research has continued ever since. Kottas *et al.* proposed a method to report confidence intervals for AUC [3]. Yu *et al.* proposed a modified AUC which is customized for gene ranking [4]. Yu also proposed another version of AUC for penalizing regression models used for gene selection with high dimensional data [5]. Rosenfeld *et al.* used AUC as a loss function and demonstrated AUC-based training lead to better generalization [6]. Their research, however, is not in the context of Neural Networks (NN); instead, they use Support Vector Machines (SVM). Therefore, their method does not address the challenges we address in this paper, including taking confidence of the model into account in calculating AUC and thus, making it a better metric for training neural networks. Zhao *et al.* proposed an algorithm for AUC maximization in online learning [7]. A stochastic approach for the same task was introduced by Ying *et al.* [8]. Cortes and Mohri studied correlation of AUC, as it is optimized, and error rate [9]. Their research showed that minimizing the error rate may not result in maximizing AUC. Ghanbari and Scheinberg directly optimized error rate and AUC of the classifiers; however, their approach only applies to linear classifiers [10].

This paper explains in detail the meaning of AUC, how reliable it is, under which circumstances it should be used, and its limitations. It also proposes a novel approach to eliminate these limitations. Our primary focus is on deep learning and Convolutional Neural Networks (CNNs), which differentiates our work from the previous work in the literature. We propose confidence-incorporated AUC (cAUC) as a modified AUC which directly correlates to Cross-Entropy Loss function and thus, helps to stop CNN training at a more optimum point in terms of confidence. This is not possible with conventional AUC, as not only the minimum of Binary Cross-Entropy loss function may not correlate with the maximum of AUC, but also AUC does not take the confidence of the model into account. We have also published a new library called GeuineAI[1], which contains our modified AUC and conventional AUC with more features in comparison to the existing standard python libraries.

## Revisiting the Concept of AUC

In supervised binary classification, each datapoint has a label. Conformed with standards of Machine Learning, labels are either 0/1 or 01/10 or sometimes +1/-1 and the model's (classifier's) outputs are usually probabilities. In the case of cancer detection, for example, input data may be CT or MRI images. Cancerous cases will be images labeled with 1 (positive) and normal (healthy) images will have 0 (negative) as their labels. The model returns a probability for each image. In the ideal scenario, the model's output will be 1 for cancerous images and 0 for normal ones.

Four possible outcomes of binary classification are True Positive (TP), True Negative (TN), False Positive (FP), and False Negative (FN). From the Table 1, it can be inferred that TX means Truly predicted as X and FX means Falsely predicted as X.

**Table 1.** Possible outcomes of binary classification

|    | Actual Value | Predicted Value |
|----|--------------|-----------------|
| TN | 0            | 0               |
| FP | 0            | 1               |
| FN | 1            | 0               |
| TP | 1            | 1               |

Defined as the total number of correct predictions out of total cases, Accuracy is calculated by Equation (1).

$$Accuracy = \frac{TP + TN}{TP + FP + TN + FN} \tag{1}$$

As it can be seen, accuracy is only concerned about correct versus wrong predictions. In many situations, especially in Medical Science, this is not enough. The consequences of misclassifying a normal case as cancerous and considering a cancerous case as normal are way different. The first one is referred to FP, also known as Type I error, whereas the second one is a FN or Type II error. True Positive Rate (TPR) and False Positive Rate (FPR) are two criterions which distinguish the error types.

$$TPR = \frac{TP}{TP + FN} \tag{2}$$

$$FPR = \frac{FP}{FP + TN} \tag{3}$$

TPR is also known as sensitivity and refers to the ratio of correct predictions to total within actual positives. FPR is the ratio of wrong predictions within actual negatives. FPR is related to specificity by Equation (4), which is used frequently in Medical Science.

$$FPR = 1 - specificity \tag{4}$$

---

[1] https://pypi.org/project/GenuineAI/

As mentioned before, predicted value should be binary, but output of the model is probability. Thresholding is how probabilities are converted to predicted values. As an example, if the output is 0.6 and the threshold is 0.5, predicted value is 1.

$$y = \begin{cases} 0 \; if \; p \leq t \\ 1, otherwise \end{cases} \tag{5}$$

y in Equation (5) is the predicted value, $p$ is the output of the model, which is a probability, and $t$ is the threshold. Depending on $t$, TPR and FPR will be different. ROC is the curve formed by plotting TPR versus FPR for all possible thresholds and AUC is the area under that curve.

In the following, we take an example-based approach to highlight the fundamentals of AUC.

**Example 1:** Table 2 contains the simplest possible example. It should be followed from left to right. $y^d$ refers to the desired value which is the same as the label (ground truth).

**Table 2** Example 1

| $y^d = 1$ | $t < 0.5$ | $y = 1$ | $TP = 1$ | $TPR = \dfrac{TP}{TP + FN} = \dfrac{1}{1 + 0} = 1$ |
| | | | $TN = 0$ | |
| $p = 0.5$ | | | $FP = 0$ | $FPR = \dfrac{FP}{FP + TN} = \dfrac{0}{0 + 0} = NaN$ |
| | | | $FN = 0$ | |

It can be seen from table 2 that actual positives and actual negatives are necessary to draw an ROC curve. Although it may seem trivial, lack of one category in one batch leads to NaN in training of Machine Learning (ML) models. Furthermore, if the batch size is equal to one, the batch AUC is always NaN. Consequently, for any NN to be directly trained with a modified AUC, or for any code where AUC is calculated within each batch, batch size of one cannot be used. Furthermore, the sampler should be customized in a way to return samples from both classes in each batch.

**Example 2:** Table 3 contains an example of classifying one positive and one negative cases and Fig. 1 shows the corresponding ROC curve. There are important points in this example. ROC curves always start from (0,0) and always end at (1,1). The reason is that if threshold is 0, all predicted values are 1. They will be either TP or FP. Therefore, both TPR and FPR are 1. On the other hand, if threshold is 1, everything is predicted as negative. In this case, predictions are all TN or FN. Consequently, TPR and FPR will be both zero. Two things must be taken into account when writing a ML code: $t=0$, and $t=1$ should be treated separately and $t$ should be iterated backward if going from $(0, 0)$ to $(1, 1)$ is desired. Backward iteration necessity comes from the fact that the highest t corresponds to the lowest TPR and FPR. Exceptions of $t=0$ and $t=1$ are needed for rare cases when the output of the model is exactly 0 or 1.

**Table 3** Example 2

| $y_1^d = 1$ $y_2^d = 0$ | $t < 0.4$ | $y_1 = 1$ $y_2 = 1$ | $TP = 1$ | $TPR = \dfrac{1}{1 + 0} = 1$ |
| | | | $TN = 0$ | |
| | | | $FP = 1$ | $FPR = \dfrac{1}{1 + 0} = 1$ |
| | | | $FN = 0$ | |
| | $0.4 \leq t < 0.6$ | $y_1 = 0$ $y_2 = 1$ | $TP = 0$ | $TPR = \dfrac{0}{0 + 1} = 0$ |
| | | | $TN = 0$ | |
| $p_1 = 0.4$ $p_2 = 0.6$ | | | $FP = 1$ | $FPR = \dfrac{1}{1 + 0} = 1$ |
| | | | $FN = 1$ | |
| | $0.6 \leq t$ | $y_1 = 0$ $y_2 = 0$ | $TP = 0$ | $TPR = \dfrac{0}{0 + 1} = 0$ |
| | | | $TN = 1$ | |
| | | | $FP = 0$ | $FPR = \dfrac{0}{0 + 1} = 0$ |
| | | | $FN = 1$ | |

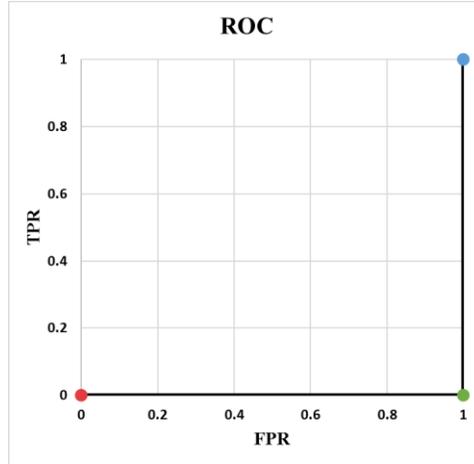

**Fig. 1.** ROC of Example 2

**Example 3:** Our third example is complement of Example 2. As it is indicated in Table 4, output probability for the positive case ($y_1{}^d$) is higher. Under these conditions, AUC is equal to 1, as depicted in Fig 2. In other words, ideal situation for classification of one positive and one negative example in terms of AUC is when output probability of the positive case is higher.

**Table 4** Example 3

| $y_1{}^d = 1$ $y_2{}^d = 0$ | $t < 0.4$ | $y_1 = 1$ $y_2 = 1$ | $TP = 1$ | $TPR = \dfrac{1}{1+0} = 1$ |
| | | | $TN = 0$ | |
| | | | $FP = 1$ | $FPR = \dfrac{1}{1+0} = 1$ |
| | | | $FN = 0$ | |
| | $0.4 \leq t < 0.6$ | $y_1 = 1$ $y_2 = 0$ | $TP = 1$ | $TPR = \dfrac{1}{1+0} = 1$ |
| | | | $TN = 1$ | |
| $p_1 = 0.6$ $p_2 = 0.4$ | | | $FP = 0$ | $FPR = \dfrac{0}{0+1} = 0$ |
| | | | $FN = 0$ | |
| | $0.6 \leq t$ | $y_1 = 0$ $y_2 = 0$ | $TP = 0$ | $TPR = \dfrac{0}{0+1} = 0$ |
| | | | $TN = 1$ | |
| | | | $FP = 0$ | $FPR = \dfrac{0}{0+1} = 0$ |
| | | | $FN = 1$ | |

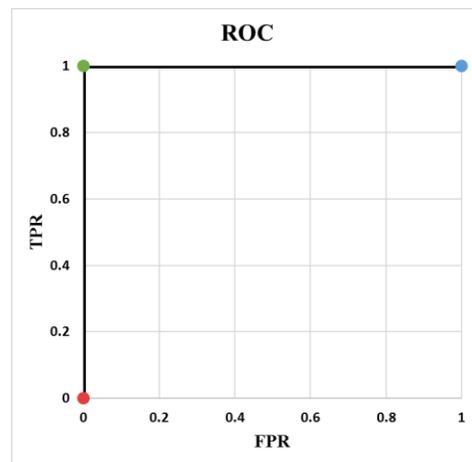

**Fig. 2.** ROC of Example 3

It should be noted that if the two probabilities were slightly different, e.g., $p_1 = 0.501$ and $p_2 = 0.499$, AUC would be 1. The separation of probabilities does not have to be at 0.5. $p_1 = 0.0002$ and $p_2 = 0.0001$ would still result in AUC = 1. This leads to an important issue which is confidence. It turns out AUC does not take into account the confidence of the model.

**Example 4:** In the fourth example (Table 5), the output probabilities are the same for the two samples. This leads to AUC of 0.50. This example shows that whenever all output probabilities are equal, AUC is 0.50 and ROC is a straight line from (0, 0) to (1, 1) (Fig 3). This is true for all different values of N where N is batch size or number of samples.

**Table 5** Example 4

| $y_1{}^d = 1$ $y_2{}^d = 0$ | $t < p$ | $y_1 = 1$ $y_2 = 1$ | $TP = 1$ | $TPR = \dfrac{1}{1+0} = 1$ |
| | | | $TN = 0$ | |
| | | | $FP = 1$ | $FPR = \dfrac{1}{1+0} = 1$ |
| | | | $FN = 0$ | |
| $p_1 = p_2 = p$ | $p \le t$ | $y_1 = 0$ $y_2 = 0$ | $TP = 0$ | $TPR = \dfrac{0}{0+1} = 0$ |
| | | | $TN = 1$ | |
| | | | $FP = 0$ | $FPR = \dfrac{0}{0+1} = 0$ |
| | | | $FN = 1$ | |

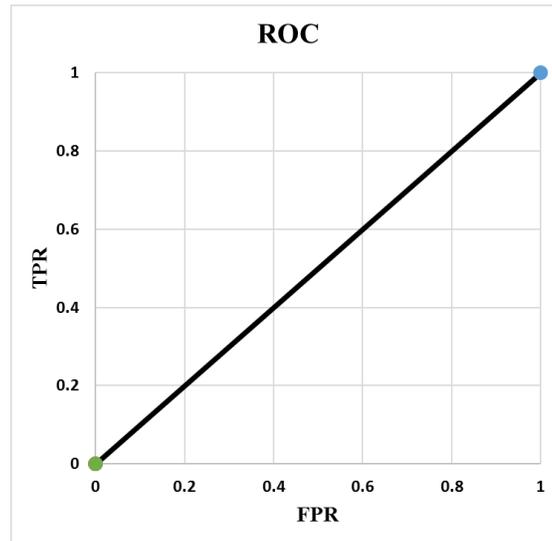

**Fig. 3** ROC of Example 4

**Example 5:** In example 5, N is equal to 3 and there are 4 points in the ROC curve. The reason for this phenomenon is effective threshold boundaries. As it can be seen in Table 6, up to t=0.4, no value of t changes the model's predictions. It turns out that those effective boundaries are defined by predicted probabilities. It should now be highlighted, in Examples 2 and 3, N was 2 and there were 3 points on the ROC curve. In the general form, for N predictions, there will be N+1 points on the ROC curve. For each pair of predictions with equal probabilities, one point is omitted. The extreme case is when all output probabilities are equal. In this case, there will be two points on the ROC curve and AUC is 0.5 (Example 4).

**Table 6.** Example 5

| | | | | | |
|---|---|---|---|---|---|
| $y_1{}^d = 0$<br>$y_2{}^d = 0$<br>$y_3{}^d = 1$ | $t < 0.4$ | $y_1 = 1$<br>$y_2 = 1$<br>$y_3 = 1$ | $TP = 1$<br>$TN = 0$<br>$FP = 2$<br>$FN = 0$ | $TPR = \dfrac{1}{1+0} = 1$<br>$FPR = \dfrac{2}{2+0} = 1$ | |
| | $0.4 \leq t < 0.45$ | $y_1 = 0$<br>$y_2 = 1$<br>$y_3 = 1$ | $TP = 1$<br>$TN = 1$<br>$FP = 1$<br>$FN = 0$ | $TPR = \dfrac{1}{1+0} = 1$<br>$FPR = \dfrac{1}{1+1} = 0.5$ | |
| $p_1 = 0.4$<br>$p_2 = 0.55$<br>$p_3 = 0.45$ | $0.45 \leq t < 0.55$ | $y_1 = 0$<br>$y_2 = 1$<br>$y_3 = 0$ | $TP = 0$<br>$TN = 1$<br>$FP = 1$<br>$FN = 1$ | $TPR = \dfrac{0}{0+1} = 0$<br>$FPR = \dfrac{1}{1+1} = 0.5$ | |
| | $0.55 \leq t$ | $y_1 = 0$<br>$y_2 = 0$<br>$y_3 = 0$ | $TP = 0$<br>$TN = 2$<br>$FP = 0$<br>$FN = 1$ | $TPR = \dfrac{0}{0+1} = 0$<br>$FPR = \dfrac{0}{0+2} = 0$ | |

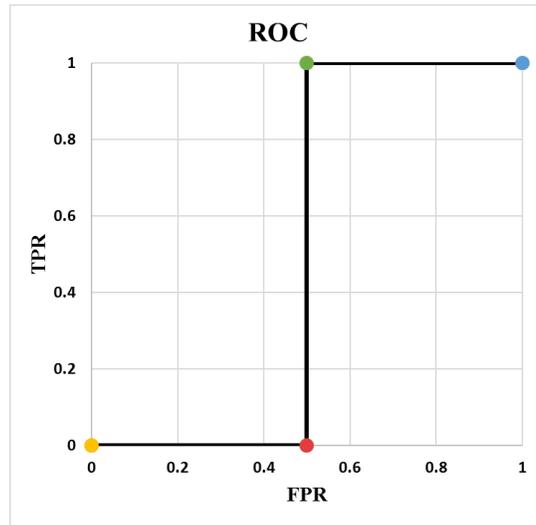

**Fig. 4.** ROC of Example 5

## Methods

Inspired by the previous examples, we will now investigate some characteristics of ROC and AUC. We will demonstrate how misclassification of a single data point can decrease AUC, and what extreme scenarios of misclassification look like. We will then provide an example to show a higher AUC does not necessarily correspond to better classification. The section is concluded with introducing cAUC, our proposed modified AUC, and mathematical support for its correlation to Binary Cross Entropy (BCE).

A result of having *N+1* points on the ROC curve is that *N+1* different effective values can be assigned to threshold *t*. In other words, while infinite values for *t* can be selected, selecting more than *N+1* values for *t* would not help to achieve more accurate AUC or "smoother" ROC curve. Even if calculations are precise, the efficiency will be degraded because if t values are not selected from different effective intervals, they will result in the same point on ROC. In Example 3, *t*=0, 0.1, 0.2, 0.3, or any other value below 0.4 will result in (1, 1) on ROC. Furthermore, because continuous variables have to be discretized, selecting fixed step size to increase $t$ may result in inaccuracy. It happens almost certainly if two probabilities are highly close to each other and the fixed step is not small enough to land between them. Usually high values of N create such circumstances. Therefore, having a method for selecting optimal threshold is crucial. Changing value of $t$ is effective if and only if it affects predictions. Assuming probabilities are sorted, any value of $t$ between $p_i$ and $p_{i+1}$ does not change predictions. Supported by the same rational, the optimal values of $t$ we suggest is given by (6). An optimum set, based on the rule of having $N+1$ points in ROC, has to have $N+1$ members. However, our proposed set has $N+2$ elements. If Equation (5) is conformed, 1 can be removed from the set. Nevertheless, adding 0 and 1 to the set is a safe approach for avoiding programming errors.

$$t \in \{\, 0, \, p_i \,, 1 \,\} \,, i = 1,2,\dots,N \tag{6}$$

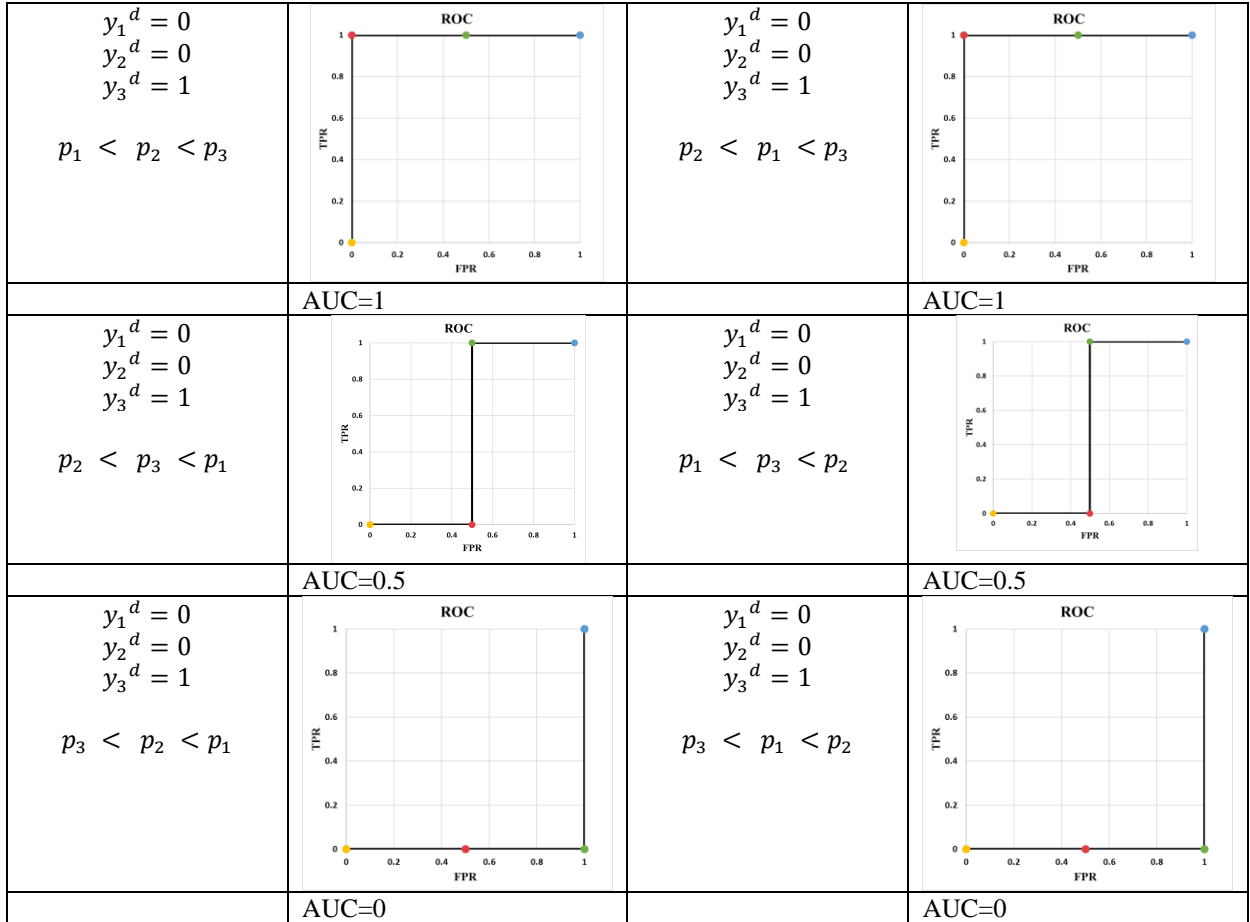

**Fig. 5** ROC curves for N=3, two actual negative and an actual positive

Fig 5 depicts all possible outcomes (except special cases of equal probabilities). It seems ROC is always staircase looking, except for the situations where a pair of predicted probabilities are equal. Thus, using trapezoid integration is the best and most accurate technique to calculate AUC. Furthermore, Fig 5 demonstrates order of predicted probabilities plays a key role in amount of AUC. If there is at least one threshold $t$ where the probabilities of all actual positives and negatives are above and below it, respectively, then the AUC is equal to 1. Although the mathematical proof needs more fundamentals, there is one key support: selecting t at the boundary of positive and negative data points results in a perfect classification corresponding to (0, 1) on ROC.

$$AUC = 1 \; if \; \exists \, t \mid \{\forall p_i, y_i^d \in AP \;\; t < p_i \; and \; \forall p_j, y_j^d \in AN \;\; p_j \leq t\} \tag{7}$$

Where *AP* and *AN* are actual positives and actual negatives, respectively.

To be able to separate positive and negative datapoints in a way that probabilities of positive cases are higher, we introduce $\varepsilon$. In Table 7, $\varepsilon$ is a positive real number which is less than or equal to $p$. This ensures $p$-$\varepsilon$ is zero or positive and implies that $p$-$\varepsilon$ is less than $p$. For example, if $p$ is 0.8, $\varepsilon$ can be in the range of 0 to 0.8. It also explains why (7) is true. For any $t \in [p - \varepsilon, p)$, conditions of (7) are met and the AUC is equal to 1. In this case, (0,0), (0,1), and (1,1) are points of ROC.

**Table 7.** A group of realizations with N=3, AN=2, and AP=1

| | t | | |
|---|---|---|---|
| Sorted Actual Values | 0 | 0 | 1 |
| Predicted Probabilities | p-ε | p-ε | p |
| | TN | TN | TP |

In Table 7, probabilities of all actual negatives are equal ($p$-$\varepsilon$). To be able to sort probabilities within each class of datapoints, $\delta$ is introduced. Table 8 extends Table 7 scenario to more general cases where probabilities are not necessarily equal. In this case, $\delta$ can be considered as a random noise which is a non-negative real number. It helps to simulate predicted probabilities better. With $\delta$, the predicted probabilities do not follow a distinct pattern of having a fixed distance.

**Table 8.** A group of realizations with N=8, AN=4, AP=4, and AUC=1

| | | | t | | | | |
|---|---|---|---|---|---|---|---|
| Sorted Actual Values | 0 | 0 | 0 | 0 | 1 | 1 | 1 | 1 |
| Sorted Probabilities | p-ε-3δ | p-ε-2δ | p-ε-δ | p-ε | p | p+δ | p+2δ | p+3δ |
| | TN | TN | TN | TN | TP | TP | TP | TP |

Table 9 shows the other extreme. When there is threshold $t$ such that probabilities of all actual positives and negatives are below and above it, respectively, then the AUC is zero.

$$AUC = 0 \ if \ \exists \ t \mid \{\forall p_i, y_i{}^d \in AP \ \ p_i \le t \ and \ \forall p_j, y_j{}^d \in AN \ \ t < p_j\} \qquad (8)$$

**Table 9.** A group of realizations with N=8, AN=4, AP=4, and AUC=0

| | | | t | | | | |
|---|---|---|---|---|---|---|---|
| sorted Actual values | 0 | 0 | 0 | 0 | 1 | 1 | 1 | 1 |
| sorted Probabilities | p-3δ | p-2δ | p-δ | p | p-ε | p-ε+δ | p-ε+2δ | p-ε+3δ |
| | FP | FP | FP | FP | FN | FN | FN | FN |

Table 10 depicts all remaining possible scenarios where AUC is greater than zero ($0 < AUC$). Table 10(a) gives the big picture. For $t \in [p - \varepsilon, \ p)$, there will be one FP in predicted values (Table 10(b)), which means TPR is 1 and FPR is positive. For $t \in [p, p + \delta)$, there will be one FN in predicted values (Table 10(c)), which means FPR=0 and TPR less than one. In other words, in the ROC curve, Table10(b) and 10(c) correspond to points $(d_1, 1)$ and $(0, d_2)$, respectively, where $d_1$ and $d_2$ are positive real numbers (Fig. 6). Obviously, this causes a reduction in AUC as much as the area of a triangle. $(0, d_2)$, $(d_1, 1)$, and $(0, 1)$ are vertices of the triangle. FN contributes to TPR whereas FP is part of FPR. Therefore, $d_1$ is influenced by FP and $d_2$, is a function of FN. Because they both play a role in the triangle's area, it can be concluded that the AUC does not discriminate between FP and FN. All it does is scaling the importance with respect to degree of imbalance. In other words, AUC equalizes importance of positive and negative cases as if the number of *APs* and *ANs* were the same. In this perspective, ROC has a built-in normalizer mechanism. However, in real world, that may not be desired. In most cancer detection situations, for example, importance of a positive case massively outweighs that of a negative case.

**Table 10.** A group of realizations with N=8, AN=4, AP=4, and 0<AUC<1

| (a) | | | | | t | | | | |
|---|---|---|---|---|---|---|---|---|---|
| | sorted Actual values | 0 | 0 | 0 | 0 | 1 | 1 | 1 | 1 |
| | sorted Probabilities | p-ε-2δ | p-ε-δ | p-ε | p | p | p+δ | p+2δ | p+3δ |
| | | TN | TN | TN | FP | FP | TP | TP | TP |
| (b) | | | | | t | | | | |
| | sorted Actual values | 0 | 0 | 0 | 0 | 1 | 1 | 1 | 1 |
| | sorted Probabilities | p-ε-2δ | p-ε-δ | p-ε | p | p | p+δ | p+2δ | p+3δ |
| | | TN | TN | TN | FP | TP | TP | TP | TP |
| (c) | | | | | | t | | | |
| | sorted Actual values | 0 | 0 | 0 | 0 | 1 | 1 | 1 | 1 |
| | sorted Probabilities | p-ε-2δ | p-ε-δ | p-ε | p | p | p+δ | p+2δ | p+3δ |
| | | TN | TN | TN | TN | FN | TP | TP | TP |

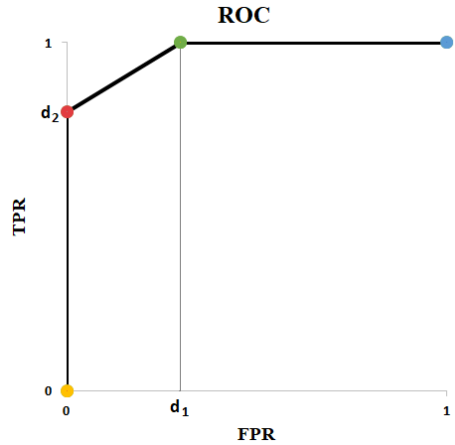

**Fig. 6.** ROC of Example of Table 10

The fact that the AUC does not discriminate between FP and FN implies that what should be used as a criterion when training a model is ROC curve itself and not the AUC. Hence, in order to translate probabilities to predictions, one specific $t \in [0, 1]$ is needed.

In medical science (e.g., cancer detection), instead of AUC value, the clinical value of a classification method is usually studied in terms of TPR or FPR. For example, for a desired TPR, using the ROC curve, the point with lowest FPR is selected. From there, the desired threshold is derived, and the classification is performed. Thus, to evaluate the performance, confusion matrix is the most informative way of reporting where a model with a lower AUC may be preferred when the specific TPR/FPR are considered. One possible example is illustrated in Fig 7.

In Fig 7, AUC of the orange line and the blue line are 0.75 and 0.58, respectively. Although the orange line has a higher AUC, if the acceptable sensitivity is set at 1, the blue line corresponds to the best model. In other words, to be able to identify every single positive example, with the orange line we will misclassify 75% of our negative examples compared with 50% of misclassification by the blue one.

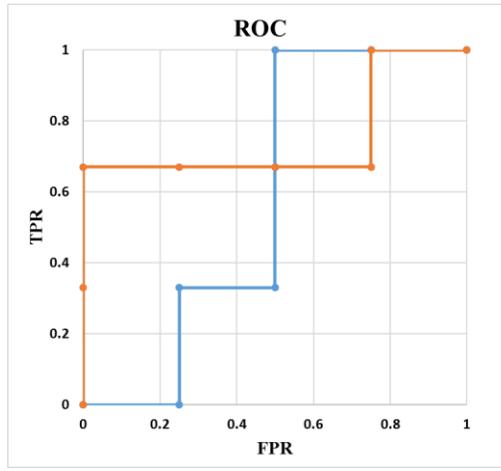

**Fig. 7.** ROC curves for two different models with N=7

**Proposed AUC with Confidence**

We call a model confident if it returns probabilities near 1 for all positive cases and probabilities near 0 for all negative examples. In previous section, it was demonstrated that AUC does not provide the confidence of the classification model under study. In other words, whether the predicted probabilities are close to each other or not does not affect the AUC value. As a result, a classification model that is able to separate the positive and negative cases by a small margin (e.g., 5%), has the same AUC as the one that separates the positive and negative cases by a large margin (e.g., 25%). Risk assessment in Medical Science and regression in Statistics are cases where having large margins may not be the target. However, in the context of classification, the margin is a key point. The whole idea of Support Vector Machines (SVM) is formed around large margin classification [11]. The ultimate effect of Cross Entropy (CE) loss function on NNs is imposing separation between predicted probability of positive and negative examples [12].

To address this issue, we propose a modified AUC (cAUC), which provides a confidence measure for the classification model. To do so, we introduce two coefficients, $\alpha$ and $\beta$.

$$\alpha = \max(p_i) - \min(p_j) \mid \{p_i \in AP, p_j \in AN\} \qquad (9)$$

$$\beta = \min(p_i) - \max(p_j) \mid \{p_i \in AP, p_j \in AN\} \qquad (10)$$

$$cAUC = e^{(\alpha-1)}e^{(\beta-1)}AUC \tag{11}$$

The idea behind Equation (11) is the smaller the range between the probabilities of the two classes, the lower the AUC will be and vice versa. If the range is the maximum possible value (which is 1), the AUC remains unchanged. Otherwise, it is decreased.

In the following, we show that our cAUC local maximums correspond to BCE local minimums. Intuitively, BCE is minimized when the probabilities created by the model are close to 1 for APs and near 0 for ANs. This translates to the concept of confidence we discussed above. Mathematically, BCE is explained through (12). Using the same separation approach, we have used so far, BCE can be rewritten for APs and ANs as (13). From (13), it can be concluded ideal BCE loss is resulted under conditions of (14).

$$BCE = \frac{-1}{N}\sum_{i=1}^{N} y^d{}_i \log(p_i) + (1 - y^d{}_i) \log(1 - p_i) \tag{12}$$

$$BCE = \frac{-1}{N}\left[\sum_{i=1}^{N}\begin{Bmatrix}\log(p_i) \mid y^d{}_i \in AP \\ 0 \quad,otherwise\end{Bmatrix} + \sum_{j=1}^{N}\begin{Bmatrix}\log(1 - p_j) \mid y^d{}_j \in AN \\ 0 \quad,otherwise\end{Bmatrix}\right] \tag{13}$$

$$BCE = 0 \; if \; \forall y_i{}^d \in AP, p_i = 1 \; and \; \forall y_i{}^d \in AN, p_i = 0 \tag{14}$$

If conditions of (14) are met, from (7) it can be inferred AUC is equal to 1 because for any threshold between 0 and 1, all datapoints are correctly classified. In this case, (9) and (10) result in $\alpha = \beta = 1$. Ultimately, our definition of cAUC, (11), returns $cAUC = 1$. Therefore, the ideal cases of cAUC and BCE correspond to each other. Through a similar procedure, it can be proved their worst cases (cAUC = 0 and BCE → ∞ ) correspond too. In the transition between the two extremes, BCE and confidence-related part of cAUC (the exponential coefficients) have a monotonic behavior.

We proved that if AUC is equal to 1, the probability of positive and negative examples can be close to each other and thus, leading to high BCE. Therefore, a high AUC does not necessarily mean low BCE. Thus, instead of AUC, we propose monitoring cAUC, which in global optimums is guaranteed to result in ideal BCE and AUC, and in local optimums has higher potential for stopping the training when the model is confident, not overfit, and achieves a high AUC.

## Results

We will evaluate our confidence-incorporated AUC (cAUC) on 4 different scenarios: random predictions, a customized dataset based on MNIST [13], our proprietary Prostate Cancer (PCa) dataset, and a dataset based on BraTS19[14][15][16]. Our PCa dataset of Diffusion-weighted MRI is described in our previous research [17]. The CNN architectures and the utilized settings are similar to our shallow models used in other research projects [18]. Nonetheless, the details are provided in Appendix A. Given the fact that AUC is not differentiable, to train the network we used BCE. The only essential point which should be covered is input channels of our CNN for MNIST classification. Because MNIST is a single channel dataset, we revised the network to be compatible with it.

**cAUC vs AUC on Random Data**

To test the proposed AUC, in an N=10 simulation, real values and predicted probabilities were generated randomly using U[0, 1] as Table 11. In case of arbitrary classification, expected value of AUC is 0.5. The goal here is to calculate expected values of cAUC for such conditions. Another point for the presented values in Table 11 is to highlight importance of sample size. With the widespread use of AI in Medical Science, researchers must care about sample sizes. Our experiment shows AUC = 0.66 is not hard to achieve through chance when N is not high enough.

**Table 11.** Comparison of AUC and the proposed AUC for a random case.

| Real Values | Sorted Probabilities | Parameters |
|:---:|:---:|:---|
| 1 | 0.803258838 | $\alpha = 0.80325884 - 0.27759354 = 0.5256653$ |
| 0 | 0.517853202 | $\beta = 0.30374599 - 0.69960646 = -0.39586047$ |
| 1 | 0.639592674 | $AUC = 0.6666666666666666$ |
| 1 | 0.303745995 | $cAUC = 0.1027290563696407$ |
| 0 | 0.699606458 | |
| 0 | 0.318090495 | |
| 0 | 0.277593543 | |
| 1 | 0.421482502 | |
| 1 | 0.556011119 | |
| 1 | 0.548716153 | |

Simulations with N=100 and 10000 trials show expected value of AUC is 0.50 and expected value of the revised AUC is 0.07. Intuitively, AUC = 0.5 happens when everything is by chance. We showed one example is when output of the model is constant. In other words, when variance of the output vector is zero. In this case, coefficients $\alpha$ and $\beta$ also are zero in limit (according to (9) and (10)). Therefore, cAUC will be $0.5 * e^{(-1)}e^{(-1)}$ which is 0.07.

**cAUC vs AUC on an MNIST-based Dataset**

MNIST is a well-known dataset of handwritten digits, including 60000 train and 10000 test images [13]. It includes single channel, 28x28 pixel, normalized images. The 10 different digits form classes of data in MNIST, by default. Because our ultimate goal was Medical applications, we marked examples of 7 as positive and all other digits as negative to create our imbalanced binary MNIST-based dataset. Our train set included the first 5000 examples of training cohort of MNIST and our validation set was 1500 examples (indices: 45000-46500) of it. Our test set was built from the first 1000 examples of MNIST test. This was done to ensure our dataset size is reasonable in comparison to Medical ones. To make our data noisy, as it is always seen in Medical datasets, we added uniform random noise to each pixel. For that end, we first scaled MNIST examples in order to have each pixel values in the range of [0, 1]. Then we added 5 times of a random image to it and scaled the result back to [0, 1] as stated in (15).

$$image = \frac{\frac{MNIST\ image}{255} + 5 * numpy.random.random((28,28))}{6} \qquad (15)$$

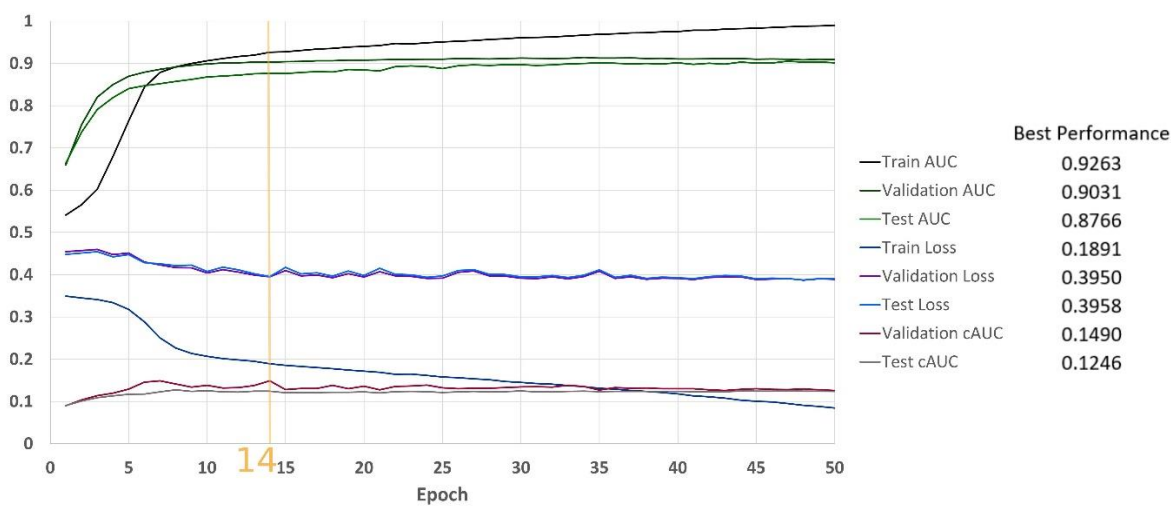

**Fig. 8.** Classification results on the MINIST-based dataset

Fig 8 shows results of the classification over 50 epochs of training. In each epoch, average BCE loss, AUC, and cAUC for training, validation, and test cohorts are calculated. This procedure is maintained until the last epoch and then the monitored values are plotted.

**cAUC vs AUC on a proprietary PCa Dataset**

Fig 9 depicts the results of classification over our institutional review board approved PCa dataset, which included Diffusion-weighted MRI images of 414 prostate cancer patients (5,706 2D slices). The dataset was divided into training (217 patients, 2,955 slices), validation (102 patients, 1,417 slices), and test sets (95 patients, 1,334 slices). Label for each slice was generated based on the targeted biopsy results where a clinically significant prostate cancer (Gleason score>6) was considered a positive label. The golden vertical line is where cAUC guides us to stop and the grey vertical line is where we would stop if AUC was used.

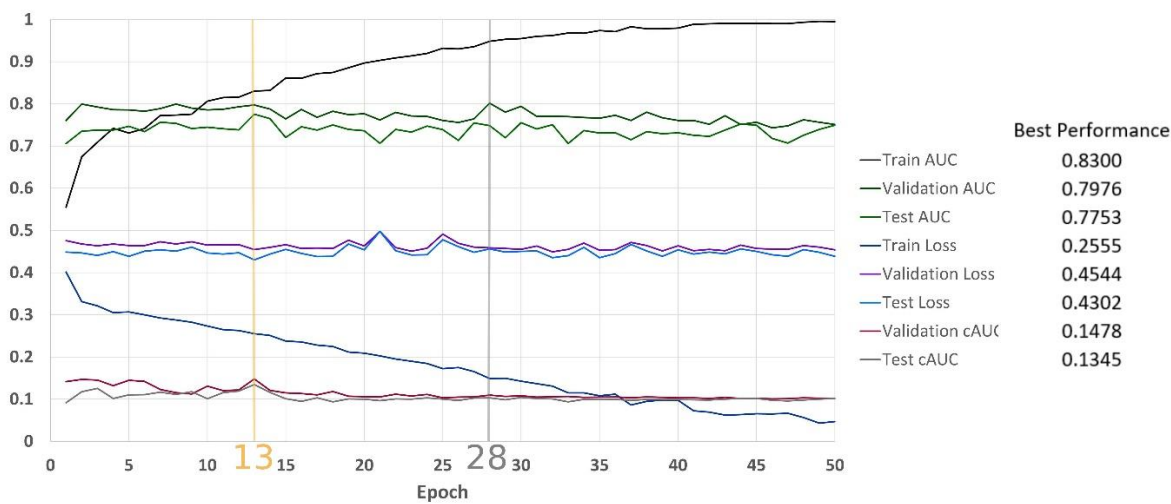

**Fig. 9.** Classification results on the PCa dataset

**cAUC vs AUC on a BraTS-based Dataset**

We used the BraTS19 dataset, with the same setting as our previous research [19]. The dataset contains 335 patients of which 259 patients were diagnosed with high-grade glioma (HGG) and 76 patients had low-grade glioma (LGG). For each patient, we stacked three MRI sequences, which are T1-weighted, post–contrast-enhanced T1-weighted (T1C), and T2-weighted (T2) volumes. With the help of BraTS segmentations, we randomly extracted 20 slices per patient with the tumor region in axial plane. Our training dataset contained 203 patients, which corresponds to 2,927 slices (1,377 LGG and 1,550 HGG examples). 66 patients were included in the validation set (970 slices, 450 LGG and 520 HGG examples). Another 66 patients formed our test set (970 slices, 450 LGG and 520 HGG examples). LGG slices were labeled as 0 and HGGs were assigned to be 1. The images were resized to $224 \times 224$ pixels. Fig 10 illustrates the results of classification over the dataset. cAUC directs the model to stop at epoch number 4 whereas both AUC and BCE would lead to the 7th epoch.

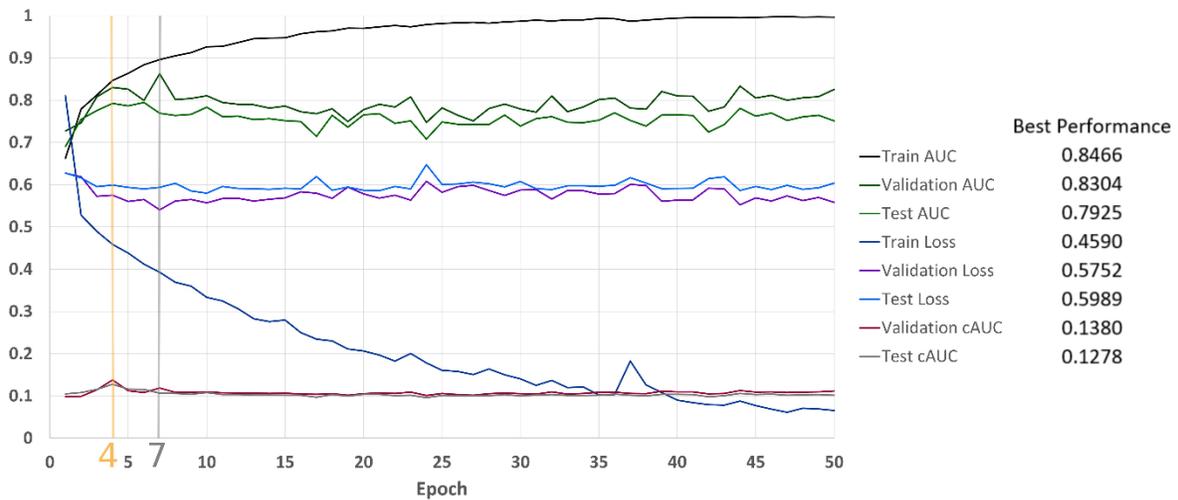

**Fig. 10.** Classification results on our BraTS-based dataset

## Discussion

In this research, we first highlighted several important ROC and AUC characteristics. We demonstrated that to draw ROC curve, both actual positives and actual negatives are needed. Threshold equal to 1 corresponds to (0,0) in the ROC curve and $t = 0$ appears as (1,1). If a function is to calculate TPR, FPR or other metrics, it should iterate backward on the $t$ values. The AUC is not concerned about confidence of the model. Regardless of N, if all the predictions are the same ($p_1 = p_2 = \cdots = p_N$), AUC will be 0.5 and the ROC curve will be a straight line from (0,0) to (1,1). Selecting more thresholds does not result in a smoother ROC or more accurate AUC. Thresholds must be selected from the set of the predicted probabilities plus 0 and 1. The order of predicted probabilities is correlated to the ROC shape and has a major impact on AUC. If there is at least a threshold where the probabilities of all actual positives and all actual negatives are above and below it, respectively, then the AUC is equal to 1. Conversely, the AUC will be 0 for the opposite case. The AUC does not differentiate FP from FN. All it does is scaling actual positive and actual negatives in a way that they have equal contributions to AUC. Therefore, the ROC curve should be used as the criterion and not AUC, if FP and FN have different weights. Because the final goal is classification, what is important is the performance of the model at a specific threshold. Therefore, there may be cases where a model with a lower AUC performs better at one threshold. The right approach is finding the optimum threshold from ROC and reporting the confusion matrix at that threshold.

The core of our research was the amendment of AUC in terms of margins. To add confidence to the optimized model, AUC needs to be refined. Using two coefficients, a revised AUC was proposed. Through simulations and mathematics, we showed the revised AUC reflects confidence of the model.

Unlike AUC, through experiments on MNIST, our PCa, and BraTS dataset, we demonstrated that local maximums in the proposed modified AUC correspond to local minimums of cross-entropy loss function. It was shown that selecting the best model based on cAUC is computationally efficient, mathematically reasonable, and it results in avoiding overfitting.

The conventional approach for when to stop training a CNN to achieve the highest AUC is to monitor the AUC while the model is being trained with a loss function such as BCE, and save the model whenever AUC breaks the previous highest score. However, when BCE is set to be used as the loss function, the hypothesis is that the best model has the lowest loss and therefore, the minimum loss is what the model is trained for. Hence, choosing the best model based on the highest AUC is not well rationalized and may not lead to the optimum point.

Our proposed metric inherits several limitations of the standard AUC and ROC but does not add any additional restrictions. Similar to AUC, cAUC is not differentiable and cannot be directly used as a loss function for training any NN. Additionally, calculating cAUC for a batch of data, especially if the batch size is small, will not help because it will be a measure of ranking in a small sample of the dataset. Similar to the standard AUC, cAUC does not give more importance to the positive examples.

## Conclusion

Our results demonstrate the proposed cAUC is a better metric to choose the best performing model. On our MNIST-based dataset, when training a CNN, it results in stopping earlier which is computationally desirable. Moreover, it has landed in a less overfitting-prone area. Our results on the prostate MRI dataset are particularly interesting. With standard AUC we would stop training the CNN model at a suboptimal point with regards to BCE. With our proposed cAUC, we are able to stop at an optimal point where the training model gives the highest AUC. Our BraTS dataset experiments demonstrate cAUC can indicate optimum points that neither AUC nor BCE would direct the model towards them.

## Appendix A: Model Architectures and Training Settings

We utilized the PyTorch version 1.9 deep learning framework to design and train the models. Tables 12-14 show details of model architectures and optimization settings for each experiment. More information abour the settings such as kernel size, stride, padding, and dilation is available in PyTorch documents[2].

**Table 12.** Architecture and settings of the MNIST experiments

| Architecture | |
|---|---|
| (cnn1): Conv2d(1, 16, kernel_size=(5, 5), stride=(1, 1)) | |
| (relu1): ReLU() | |
| (maxpool1): MaxPool2d(kernel_size=2, stride=2, padding=0, dilation=1) | |
| (cnn2): Conv2d(16, 32, kernel_size=(5, 5), stride=(1, 1)) | |
| (relu2): ReLU() | |
| (maxpool2): MaxPool2d(kernel_size=2, stride=2, padding=0, dilation=1) | |
| (fc1): Linear(in_features=512, out_features=2, bias=True) | |
| Settings | |
| Batch Size | 1 |
| Optimizer | Stochastic Gradient Descent (SGD)[20] |
| Loss function | Weighted BCE |
| Learning rate | 0.0002 |
| Momentum | 0.8 |
| L2 Penalty | 0.001 |
| Maximum number of epochs | 50 |

**Table 13.** Architecture and settings of the PCa experiments

| Architecture | |
|---|---|
| (cnn1): Conv2d(6, 16, kernel_size=(7, 7), stride=(1, 1)) | |
| (maxpool1): MaxPool2d(kernel_size=2, stride=2, padding=0, dilation=1) | |
| (relu1): ReLU() | |
| (dropout1): Dropout(p=0.1, inplace=False) | |
| (cnn2): Conv2d(16, 32, kernel_size=(5, 5), stride=(1, 1)) | |
| (maxpool2): MaxPool2d(kernel_size=2, stride=2, padding=0, dilation=1) | |
| (relu2): ReLU() | |
| (dropout2): Dropout(p=0.1, inplace=False) | |
| (cnn3): Conv2d(32, 64, kernel_size=(4, 4), stride=(1, 1)) | |
| (maxpool3): MaxPool2d(kernel_size=2, stride=2, padding=0, dilation=1) | |
| (relu3): ReLU() | |
| (dropout3): Dropout(p=0.1, inplace=False) | |
| (fc1): Linear(in_features=1024, out_features=256, bias=True) | |
| (fc2): Linear(in_features=256, out_features=64, bias=True) | |
| (fc3): Linear(in_features=64, out_features=2, bias=True) | |
| Settings | |
| Batch Size | 1 |
| Optimizer | SGD |
| Loss function | Weighted BCE |
| Learning rate | 0.0001 |
| Momentum | 0.8 |
| L2 Penalty | 0.001 |
| Maximum number of epochs | 50 |

---

[2] https://pytorch.org/docs/stable/index.html

**Table 14.** Architecture and settings of the BraTS experiments

| Architecture |  |
|---|---|
| (cnn1): Conv2d(3, 16, kernel_size=(7, 7), stride=(1, 1)) | |
| (maxpool1): MaxPool2d(kernel_size=2, stride=2, padding=0, dilation=1) | |
| (relu1): ReLU() | |
| (dropout1): Dropout(p=0.1, inplace=False) | |
| (cnn2): Conv2d(16, 32, kernel_size=(5, 5), stride=(1, 1)) | |
| (maxpool2): MaxPool2d(kernel_size=2, stride=2, padding=0, dilation=1) | |
| (relu2): ReLU() | |
| (dropout2): Dropout(p=0.1, inplace=False) | |
| (cnn3): Conv2d(32, 64, kernel_size=(4, 4), stride=(1, 1)) | |
| (maxpool3): MaxPool2d(kernel_size=2, stride=2, padding=0, dilation=1) | |
| (relu3): ReLU() | |
| (dropout3): Dropout(p=0.1, inplace=False) | |
| (fc1): Linear(in_features=36864, out_features=256, bias=True) | |
| (fc2): Linear(in_features=256, out_features=64, bias=True) | |
| (fc3): Linear(in_features=64, out_features=2, bias=True) | |
| **Settings** | |
| Batch Size | 1 |
| Optimizer | SGD |
| Loss function | Weighted BCE |
| Learning rate | 0.0001 |
| Momentum | 0.85 |
| L2 Penalty | 0.001 |
| Maximum number of epochs | 50 |

# Appendix B: AUCPlus

In python, sklearn[3] is the most renowned package containing functions for AUC and ROC. We have developed a function called AUCPlus published in GenuineAI[4] package. For the case described in Table 15, Fig 11 shows the ROC curves created by sklearn as well as AUCPlus. Sklearn uses derivatives to change values of threshold. This results in unpredictable number of points in the ROC curve. In other words, sklearn retains minimum number of points needed for drawing an ROC curve. Although in theory at each fixed TPR level, minimum FPR is desired, there are cased we may prefer to have a margin. For example, if there is a noise which imposes randomness and uncertainty, we will move away from the minimum FPR. In those cases, AUCPlus will be helpful.

---

The main limitation of sklearn is that it only returns thresholds and ROC. Our function, on the other hand, returns a dataframe containing thresholds along with TP, TN, FP, FN, TPR, FPR, Specificity, Accuracy, Precision, and F1 score at each t.

**Table 15.** Randomly generated probabilities and Real Values (N=10)

| Actual values | 0 | 0 | 0 | 0 | 1 | 1 | 1 | 1 | 0 | 1 |
|---|---|---|---|---|---|---|---|---|---|---|
| sorted Probabilities | 0.1 | 0.2 | 0.3 | 0.4 | 0.5 | 0.6 | 0.7 | 0.8 | 0.9 | 1 |

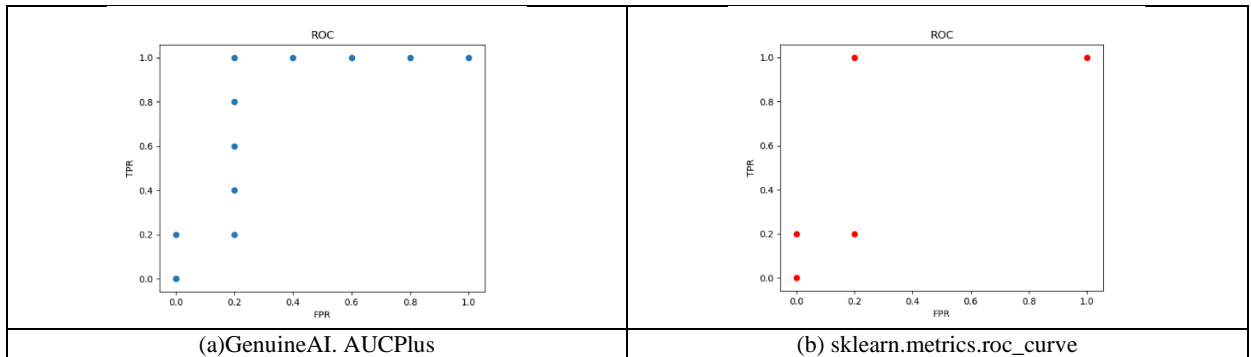

(a)GenuineAI. AUCPlus                (b) sklearn.metrics.roc_curve

**Fig. 11.** SKLearn ROC vs GenuineAI ROC (N=10)